\documentclass[letterpaper]{article} 
\usepackage{aaai24}  
\usepackage{times}  
\usepackage{helvet}  
\usepackage{courier}  
\usepackage[hyphens]{url}  
\usepackage{graphicx} 
\usepackage{natbib}  
\usepackage{caption} 
\usepackage{algorithm}
\usepackage{algorithmic}
\usepackage{csquotes}
\usepackage{graphicx}
\usepackage{amsmath}
\usepackage{amsthm}
\usepackage{amssymb}
\usepackage{multirow}
\usepackage{colortbl}
\definecolor{mygray}{gray}{.9}
\usepackage{booktabs}

\usepackage{newfloat}
\usepackage{listings}
\DeclareCaptionStyle{ruled}{labelfont=normalfont,labelsep=colon,strut=off} 
\floatstyle{ruled}
\newfloat{listing}{tb}{lst}{}
\floatname{listing}{Listing}
\title{CIDR: A Cooperative Integrated Dynamic Refining Method for Minimal Feature Removal Problem}
\author{
    Qian Chen$^1$, Taolin Zhang$^2$, Dongyang Li$^1$, Xiaofeng He$^1$$^{\text{,}}$$^3$\thanks{Corresponding author}
}
\affiliations{
    $^1$School of Computer Science and Technology, East China Normal University, Shanghai, China\\
    $^2$Alibaba Group\\
    $^3$NPPA Key Laboratory of Publishing Integration Development, ECNUP, Shanghai, China\\
    qianchen901005@gmail.com, zhangtl0519@gmail.com, dongyangli0612@gmail.com\\
    hexf@cs.ecnu.edu.cn

}

\usepackage{bibentry}

\begin{document}

\maketitle

\begin{abstract}
The minimal feature removal problem in the post-hoc explanation area aims to identify the minimal feature set (MFS). Prior studies using the greedy algorithm to calculate the minimal feature set lack the exploration of feature interactions under a monotonic assumption  which cannot be satisfied in general scenarios. In order to address the above limitations, 
we  propose a \textbf{C}ooperative \textbf{I}ntegrated \textbf{D}ynamic \textbf{R}efining method (CIDR) to efficiently  discover  minimal feature sets. Specifically, we design Cooperative Integrated Gradients (CIG) to detect interactions between features. By incorporating CIG and  characteristics of the minimal feature set, we transform the minimal feature removal problem into a knapsack problem. Additionally, we  devise an auxiliary
Minimal Feature Refinement algorithm to determine the  minimal feature set from numerous candidate sets.
To the best of our knowledge, our work is the first to address the minimal feature removal problem in the field of natural language processing. Extensive experiments demonstrate that CIDR is capable of tracing  representative minimal feature sets with improved interpretability across various models and datasets.
\end{abstract}

\section{Introduction}

Deep neural networks have achieved remarkable accomplishments in the natural language processing domain \cite{transformer,bert,gpt3,gpt4}. Meanwhile, the black-box nature of deep learning techniques has made it challenging to applications. In addition to model accuracy, there is  a demand for considering interpretability \cite{olex-etal-2019-nlp,mosca-etal-2022-suspicious}. Recently feature importance-based explanation methods have gained popularity due to fidelity and usability \cite{sikdar-etal-2021-integrated,Sekhon_Chen_Shrivastava_Wang_Ji_Qi_2023}.
\begin{figure}[!htbp]
    \centering
    \includegraphics[width=0.5\textwidth]{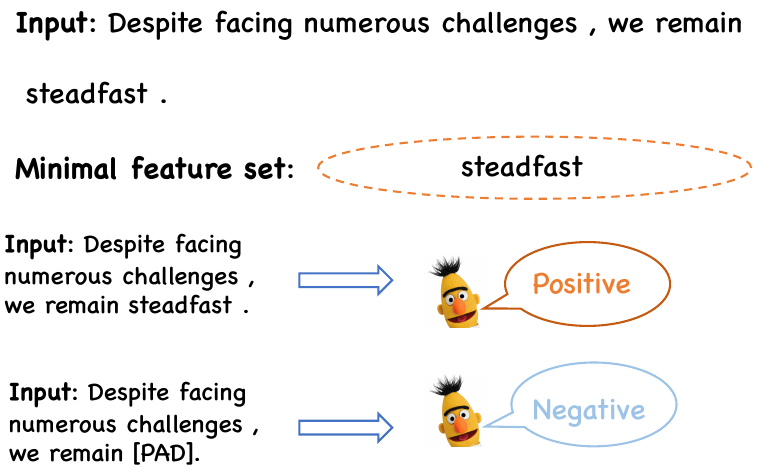}
    \caption{An illustration of the minimal feature removal problem. The bottom part shows that removing the feature of MFS would cause a drastic shift in model output probability. }
    \label{fig:introduction}
\end{figure}
For the text classification task, feature importance-based explanation methods predominantly investigate the impact of words on model prediction behavior \cite{li-etal-2016-visualizing,godin-etal-2018-explaining,Chen2020LearningVW}. Earlier researches mainly utilize the LIME algorithm \cite{lime}, attention mechanism \cite{wiegreffe-pinter-2019-attention} or gradient saliency \cite{kindermans2018learning} to ascribe a score to an individual word, measuring its influence on the output. However, such methods may lead to feature redundancy and fail to account for the synergistic interplay among words. Each word  plays a different role within a given sentence, and  word combinations matter. For example, as illustrated in Figure \ref{fig:introduction}, in the sentence \enquote{despite facing numerous challenges, we remain steadfast}, `challenges' poses a negative attitude while `remain steadfast' represents the opposite. 

Some other studies employ hierarchical clustering algorithms to detect feature interactions \cite{singh2018hierarchical,chen-etal-2020-generating-hierarchical}, coming up with low computational efficiency. \cite{Sekhon_Chen_Shrivastava_Wang_Ji_Qi_2023} focus  on using graph neural networks (GNNs) to model interactions. However, they neglect  redundant words, which may hinder the model's interpretability.
To eliminate redundant features,  \cite{harzli2023cardinalityminimal} provide a formal definition of the minimum feature removal problem proven to be an NP-complete problem and devise a greedy algorithm to solve this problem based on a monotonic assumption. However, the monotonic assumption requires non-negative model parameters, and the problem solely examines isolated feature effects, presenting obstacles to generalization. 
In this paper, we propose a \textbf{C}ooperative \textbf{I}ntegrated \textbf{D}ynamic \textbf{R}efining method, namely CIDR. Particularly, considering the  property of semantic representation through word combinations, we initially implement minor revisions to the definition of the minimal feature removal problem to rectify the inherent limitation in modeling feature interactions. Furthermore, we devise a Cooperative Integrated Gradients (CIG) formulation  to access the interaction relationship between two features by collectively computing their cooperative and individual contributions to the output probability. Based on an analogy between minimal feature sets and knapsacks,  we  utilize CIG to assign  scores for word pairs to detect feature interactions, then we regard the word pairs as items and CIG scores as corresponding weights in the knapsack problem, which is a fundamental and well-studied combinatorial optimization problem. 
In addition, due to the intractability to obtain the cardinality of the minimal feature set, we design a Minimal Feature Refinement algorithm  by introducing auxiliary variables as item values to perturb the upper bound for the maximum capacity of the `knapsacks' with producing abundant candidate sets. Eventually, we  extract the minimal features from the generated candidate sets.
The main contributions are as follows:
\begin{itemize}
\item We propose a Cooperative Integrated Dynamic Refining method (CIDR) to approach the minimal feature removal problem.  We design CIG to unearth the feature associations. By combining CIG and minimal feature properties, we transform the  original problem into a knapsack problem. Based on this finding, we devise a Minimal Feature Refinement algorithm to obtain the minimal feature set.

\item We conduct extensive experiments to verify that CIDR has the capacity to identify more interpretable minimal feature sets compared to other feature importance-based baselines. 
\end{itemize}

\section{Related Work}
There has been an increasing focus on designing interpretability algorithms to  assist in understanding the underlying principles behind a model's predictive behavior \cite{relatedwork1,relatedwork2,relatedwork3,relatedwork4,ig_v1}. Feature importance-based algorithms belong to post-hoc explanation type where input features are measured by numeric scores reflecting the significance of the features towards the prediction label. Algorithms that generate attributions can be broadly categorized into two groups: model-agnostic algorithms, such as LIME \cite{lime}, integrated gradients \cite{sundararajan2017axiomatic,ig1,ig2,DIG,enguehard-2023-sequential}, and model-dependent methods, such as LRP\cite{voita-etal-2021-analyzing}, VMASK \cite{Chen2020LearningVW} and WIGRAPH \cite{Sekhon_Chen_Shrivastava_Wang_Ji_Qi_2023}. Model-agnostic methods do not require understanding the specific details of the model and can be directly applied to any neural network, while model-dependent techniques necessitate integration with the intrinsic components of the model to augment interpretability.
IG \cite{sundararajan2017axiomatic} has stood out due to its computational efficiency and ideal explanation axioms. In recent years, there have been multiple advancements of IG in NLP domain. For example, \cite{DIG} propose DIG  to compute integrated gradients by incorporating similar words as interpolation points between two given words, SIG\cite{enguehard-2023-sequential} is designed to compute the importance of each word in a sentence while keeping all other words fixed.

Prior studies point out methods that assign attributions to features lack the construction of feature interactions \cite{NEURIPS2020_443dec30}. To address this issue,  \cite{Hao_Dong_Wei_Xu_2021} propose to integrate self-attention mechanisms to analyze the information interaction within Transformer,  \cite{chen-etal-2020-generating-hierarchical} devise a hierarchical visualization algorithm to detect feature interactions. 

Meanwhile, it is noticeable that attribution methods are highly susceptible to input perturbations, which can yield less precise attribution scores \cite{fragile,fragile3}.  To minimize the inclusion of redundant features, \cite{harzli2023cardinalityminimal} propose the minimal feature removal problem, where the minimal feature set functions as an exceedingly compact portrayal of the input features, discarding extraneous features to discern the critical components that drive the output. Additionally, it can be regarded as an attack on the output, as its definition guarantees that eliminating any feature will alter the model's output.

\section{Problem Definition}
In this section, we present the raw definition of the minimal feature removal problem and the modified version.
For the input text $x$ consisting of $n$ words, we define the collection representation of $x$ as $X=\{w_1,w_2,\cdots,w_n\}$, where $w_i$ represents the $i$-th word. Specifically, $x^{\prime}$ denotes the baseline defined in integrated gradients (IG), and $X^{\prime}$ represents its collection, which consists of $n$ [PAD] tokens. We use $S^p$ to denote the set of all word pairs. For example, suppose $X=\{w_0,w_1,w_2\}$, then $S^p=\{(w_0,w_1),(w_0,w_2),(w_1,w_2)\}$.
Given  $X$, $X^{\prime}$,
$S\subset X$ and $S^{p^{\prime}}\subset S^p$ we represent the padded input sentence collection as $X_{S}$ which is obtained from  $X$ by padding each word $w_i\in X$ where $w_i=w_j,w_j \in S$.  Similarly we represent the padded input $X_{S^p}$ which is obtained from $X$ by padding each word $w_i\in X$ where $w_i=w_j$ or $w_i=w_k,(w_j,w_k)\in S^p$. For example, suppose $S^{p^{\prime}}=\{(w_0,w_1)\}$, then $X_{S^{p^{\prime}}}=\{[\text{PAD}],[\text{PAD}],w_2,w_3\}$.

Suppose we have a function $F$ that represents a deep network and $F(X|c)$ is the output probability of given label $c$, $t$ is a positive numeric threshold ranging from $0$ to $1$. 
Original minimal feature removal problem is defined as finding a set $S_{min}\subset X$ satisfying the following two properties \cite{harzli2023cardinalityminimal}:
\begin{itemize}
    \item \textbf{Feature\enspace Essence}.\enspace$F(X_{S_{min}}|c)\leq t$.\enspace This property indicates that the features within the MFS are crucial, and their removal leads to a significant decrease in the prediction probability.
    \item \textbf{Feature\enspace
    Minimality}. Our goal is to minimize $|S_{min}|$, ensuring that for all $S^{\prime}$, which are proper subsets of $S_{min}$, $F(X_{S^{\prime}}|c) > t$. 
    This characteristic implies that $|S_{min}|$ should be sufficiently small, as the removal of any feature from $S_{min}$ would violate the feature essence. For example, without the feature minimality requirement, the set $X$ would satisfy the feature essence requirement. This is due to the clear understanding that eliminating all features would inevitably result in a reduction in the prediction probability.
\end{itemize}

On a holistic level, feature essence imposes a constraint on the MFS to prevent it from  becoming undersized, while feature minimality adds a constraint to prevent it from becoming overly large.

Apparently the original problem only explores the impact of individual features on the output. To capture feature interactions, we  slightly modify the minimal feature removal problem definition. We treat two features as one element of the MFS, then compute the MFS from all the pairwise combinations of features.  The requirements for feature essence and feature minimality remain unchanged. The modified definition is as follows:

\emph{Finding a set $S^p_{min}\subset S^p$ satisfying the feature essence and feature minimality given input $X$, $S^p$ and threshold $t$.}\footnote{In the subsequent discussions, unless otherwise specified, the term \enquote{minimal feature removal problem} refers to the modified version and the term \enquote{minimal feature set} refers to $S^p_{min}$.
}

\section{Methodology}

\begin{figure*}[!t]
    \centering
    \includegraphics[width=1.0\textwidth]{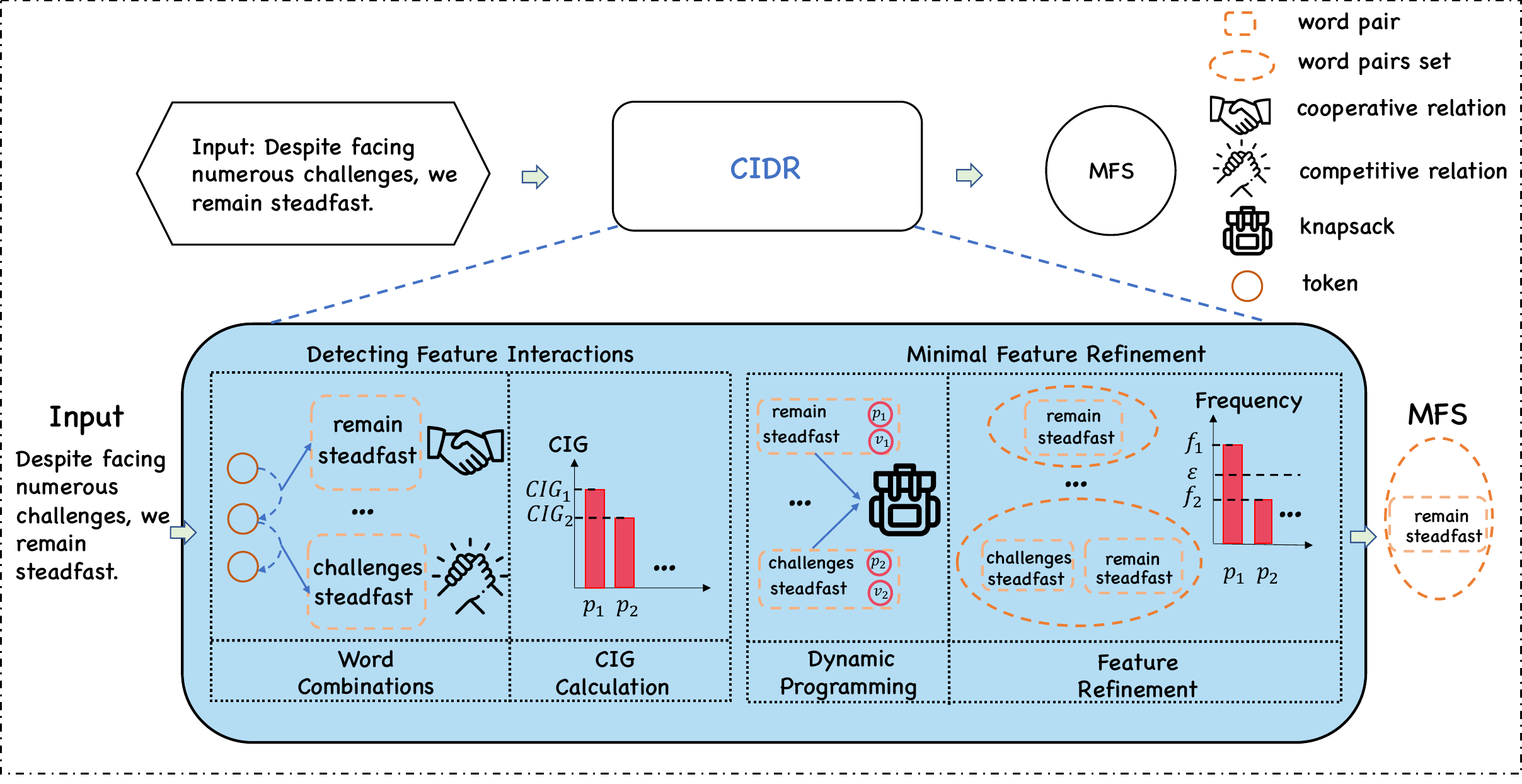}
    \caption{CIDR method (components inside the blue box): Our method workflow is illustrated at the bottom of the diagram above. Firstly, we generate word pairs by combining every two words from the input sentence. Next, we calculate the cooperative integrated gradients  (e.g. $CIG_1$,$CIG_2$) for each pair (e.g.$p_1$,$p_2$). Then, we estimate the upper bound of the minimum feature set by applying  perturbation variables (e.g. $v_1$,$v_2$) and resolve the transformed knapsack problem using a dynamic programming algorithm, resulting in multiple candidate sets. Finally, we  filter out the \enquote{false positive} minimal features by comparing the frequencies (e.g. $f_1$,$f_2$) with the threshold $\varepsilon$. }
    \label{cidr}
\end{figure*}

Inspired by \cite{sundararajan2017axiomatic}, we approach this  problem from the perspective of integrated gradients. Initially, we present a simple yet vital proposition to assist subsequent analysis.
As shown in Figure \ref{cidr}, we propose a novel approach called Cooperative Integrated Gradients (CIG) to detect feature interactions. 
Leveraging  CIG,  we approximate the minimum feature problem by transforming it into a knapsack problem. Consequently we devise the minimal feature refinement algorithm to solve the converted problem. 

For simplicity, we represent the cooperative integrated gradient of the word pair $(w_i,w_j)$ as $\text{CIG}_{i,j}$ and the MFS as $S_{min}^p$. Specially we denote the  MFS for raw definition as $S_{min}$ and the integrated gradients of word $w_i$ as $\text{IG}_i$. We define the set $X_{pos}$ as the set consisting of all words in $X$ with IG greater than 0 and denote the set containing all word pairs in $S^p$ with CIG greater than 0 as $S^p_{pos}$. 

\textbf{Proposition 1.}  $\forall w_k\in S_{min},\text{IG}_k> 0$.

\subsection{Detecting Feature Interactions with Cooperative Integrated Gradients}
Inspired by cooperative game theory \cite{Harsanyi1982,shapley1953value}, we present the solution  to model the feature interactions, namely CIG.

In order to match the original formula of IG, we replace 
$F(X|c)$ with $F(x)$ here. Given input $x$, word $w_i$ and $w_j$, we  calculate $\text{CIG}_{i,j}$ as follows:
\begin{equation}
\label{CIGeq}
\begin{split}
&\text{CIG}_{i,j}=\text{IG}_i+\text{IG}_j+\beta\left(\text{IG}_{i,X \setminus \{w_j\}}+\text{IG}_{j,X \setminus \{w_i\}}\right),
\end{split}
\end{equation}
where 
\begin{equation}
        \text{IG}_{i,X \setminus \{w_j\}} = (x_i-x_i^{\prime})\int_{\alpha=0}^1\frac{\partial F(x^{\prime}+\alpha (x^{-j}-x^{\prime}))}{\partial x_i},
\end{equation}
\begin{equation}
        \text{IG}_{j,X\setminus \{w_i\}} = (x_j-x_j^{\prime})\int_{\alpha=0}^1\frac{\partial F(x^{\prime}+\alpha (x^{-i}-x^{\prime}))}{\partial x_j},
\end{equation}
where $x^{-j}$ denotes the input $x$ with the removal of the $j$-th word. Similarly, $x^{-i}$ denotes the input $x$ with the removal of the $i$-th word, $\beta\in \mathbb{R}^+$ is the coefficient balancing the individual contributions of two features and their cooperative contributions, with values ranging from 0 to 1. Due to the inclusion of word $w_j$  in the input $x$ during the calculation of $\text{IG}_i$, we assume that $\text{IG}_i$ represents the cooperative contribution of word $w_i$. The same goes for $\text{IG}_j$.
We regard the individual contribution of two words as $\text{IG}_{i,X \setminus \{w_j\}}$ and $\text{IG}_{j,X \setminus \{w_i\}}$. Given two words, if the sum of their cooperative contributions  is greater than their individual  contributions, we consider them  engaged in a `cooperative' relationship; otherwise, they are deemed to be in a `competitive' relationship. Calculating the individual contribution is  similar to calculating the Shapley value of a feature in cooperative game theory \cite{shapley1953value}, which considers the feature's contribution to each subset of the feature set. However, due to computational complexity, we do not enumerate all subsets here but only consider two cases: $x^{-i}$ and  $x^{-j}$. $x^{-i}$ corresponds to the subset obtained by removing word $i$. 

On the basis of Proposition 1, we introduce two additional propositions specifically for CIG.

\textbf{Proposition 2.} $\forall (w_i,w_j)\in S^p_{min},\text{CIG}_{i,j}> 0$.

\textbf{Proposition 3.} $\sum_{(w_i,w_j)\in S^p_{pos}\setminus S^p_{min}} \text{CIG}_{i,j}\leq U_1+U_2$,
where 
\begin{equation}
\label{u1}
    U_1=2(|S_{pos}|-1)\sum_{w_i\in S_{pos}}\text{IG}_i,
\end{equation}

\begin{equation}
\label{u2}
    \begin{aligned}
U_2=\beta\sum_{(w_i,w_j)\in S^p_{pos}}(\text{IG}_{i,X\setminus \{w_j\}}+\text{IG}_{j,X\setminus \{w_i\}}).
    \end{aligned}   
\end{equation}
The detailed proof of Proposition 1, 2 and 3 is available in Section \ref{sec:proof}.

\subsection{Minimal Feature Refinement Algorithm}
Although \cite{harzli2023cardinalityminimal} propose  a greedy algorithm to solve the original minimal feature removal problem under the monotonicity assumption, achieving monotonicity in general neural networks is  challenging. 

In this section we leverage Proposition 2 and 3 to approximate the minimum feature removal problem as a knapsack problem.
Our core idea is to treat $S^p_{pos}\setminus S^p_{min}$ as a knapsack, where each element (i.e. word pair) in the set is considered as an item and the CIG of the element is regarded as the weight of the item. By applying Proposition 3, we can estimate the upper bound of the knapsack's capacity (i.e. $U_1+U_2$). Furthermore, based on Proposition 2 and the feature minimality, we aim to minimize $|S^p_{min}|$. In turn, this implies maximizing $|S^p_{pos}\setminus S^p_{min}|$ due to  $|S^p_{pos}|$ is constant given input $X$, $S^p$, label $c$ and the function $F$.

Unfortunately, we are unable to determine the precise  value of the maximum capacity of the knapsack.
To mitigate this, we introduce  a perturbation variable $v_{i,j}\in \mathbb{R}^+$ sampled from the standard Gaussian distribution for each word pair $(w_i,w_j)\in S^p$  to perturb the upper bound $U_1+U_2$, allowing generating different candidate sets through solving the transformed knapsack problem. We use a dynamic programming algorithm and the details are in Section \ref{sec::dynamic_programming_algorithm}.

For simplicity, we introduce the 0-1 variable $\textbf{z}_{i,j}$ to indicate
whether the word pair $(w_i,w_j)$ belongs to $S^p_{pos} \setminus S^p_{min}$,  $v_{i,j}$ is the corresponding perturbation variable.

Eventually, the minimum feature removal problem can be transformed into the following knapsack problem:

\begin{equation}
\label{problem2}
    \begin{aligned}
    &max_{\textbf{z}}\sum_{(w_i,w_j)\in S^p} v_{i,j}\textbf{z}_{i,j}\\
    s.t. &\sum_{(w_i,w_j)\in S^p} \text{CIG}_{i,j}\textbf{z}_{i,j}\leq U_1+U_2^{\prime},
    \end{aligned}
\end{equation}
where $v_{i,j} \in (0,1)$,
\begin{equation}
\label{u2}
    \begin{aligned}
U_2^{\prime}= \beta\sum_{(w_i,w_j)\in S^p_{pos}}v_{i,j}(\text{IG}_{i,X\setminus \{w_j\}}+\text{IG}_{j,X\setminus \{w_i\}}).
    \end{aligned}   
\end{equation}

In order to identify the MFS more accurately, we further propose a minimal feature refinement strategy inspired by \cite{dai-etal-2022-knowledge}. We assume that candidate sets comprise two distinct types of features: \enquote{true positive} features and \enquote{false positive} features. Our refinement strategy is designed to eliminate the minimal features that are identified as \enquote{false positive}. The key idea behind this strategy is that different candidate sets are likely to share the \enquote{true positive} minimal features, while the \enquote{false positive} ones will not. By focusing on the widely shared features, we can effectively filter out the \enquote{false positive} minimal features.

Given an input sentence, the process of identifying the MFS can be outlined as follows:
 (1) Randomly sample $v_{i,j}$ to generate $n_{iter}$ candidate sets by solving Problem \ref{problem2}.
(2) Calculate the occurrence frequency of each feature within each candidate set.
(3) Aggregate all the candidate sets, select and retain only those features shared by more than a specified threshold $t$ of the sets.

The details of the minimal feature refinement algorithm  are illustrated in Section \ref{sec::minimal_feature_refinement_algorithm}.

\section{Experiments}
 In this experiment, we  validate the effectiveness of our method by answering the following two questions:
\begin{itemize}
    \item Do the MFS derived from our method adequately satisfy the feature essence and feature minimality?
    \item Do the CIDR generate more interpretable results?
\end{itemize}

\subsection{Experimental Setups}
\noindent \textbf{Datasets}\quad To evaluate the effectiveness of CIDR, we perform comprehensive experiments on three representative binary classification datasets: \enquote{SST2} \cite{socher-etal-2013-recursive},  \enquote{IMDB} \cite{maas-etal-2011-learning}, \enquote{Rotten Tomatoes} \cite{pang-lee-2005-seeing}. 

\textbf{Dataset Statistics}\quad The statistical information for the three datasets  is shown in Table \ref{datasettable}.
\begingroup
\setlength{\tabcolsep}{4pt} 
\renewcommand{\arraystretch}{1.0}
\begin{table}[h!]
\begin{tabular}{lllll}
\hline
Dataset         & Train/Dev/Test & C & V     & L   \\
\hline
SST2            & 6920/872/1821  & 2 & 16190 & 50  \\
\hline
IMDB            & 20K/5K/25K     & 2 & 19571 & 250 \\
\hline
Rotten Tomatoes & 10K/2K/2K      & 2 & 15420 & 50 \\
\hline
\end{tabular}
\caption{Statistics of three datasets. C: number of classes, V: vocabulary size, L: average text length}
\label{datasettable}
\end{table}

\noindent \textbf{Metrics}\quad Following prior literature \cite{eraser}, we use the following two automated metrics:
\begin{itemize}
    \item \textbf{Comprehensiveness} (Comp) score \cite{AOPC} is the average difference of the change in predicted class probability before and after removing top $K$ words.  Higher is better.
    \item \textbf{Log-odds} (LO) score \cite{deeplift} is  defined as the average difference of the negative logarithmic probabilities on the predicted class before and after removing the top $K$ words. Lower is better.
\end{itemize}
In this experiment, we use LO and Comp metrics to evaluate the degree to which the obtained set satisfies the feature essence. In particular, if the set satisfies the feature essence, there should be a substantial change in probability before and after removing $K$ features of  the set. In order to align with the setting by \cite{DIG}, we remove $K$ word pairs of the set which $K$ is set to $min(0.1\times |X|,|S_{min}^p|)$. As for the feature minimality, we introduce a new metric: Feature Minimality Score (FMS). Concretely, to verify if $S^p_{min}$ satisfies the feature minimality, we calculate a removal probability, denoted as $F(X_{S^p_{min}\setminus \{pair_k\} }|c)$, for each word pair in the set. If the removal probability exceeds the threshold $t$ for all word pairs, then the set is considered to satisfy the feature minimality.  Given $N$ sequences, we compute the FMS score as follows:
\begin{equation}
    \text{FMS} = \frac{\sum_{i=1}^Nmin_{1\leq k\leq |S^p_{min}|}{\text{FM}_k^p\cdot \text{FE}^p_k}}{N},
\end{equation}
where $\text{FM}_k^p=\mathbb{I}(F(X_{S^p_{min}\setminus \{pair_k\}}|c)\thinspace\thinspace\textgreater\thinspace\thinspace t)$, $\text{FE}_k^p=\mathbb{I}(F(X_{S^p_{min}}|c)\leq t)$, 
 $pair_{k}\in S^p_{min}$. 
Similarly, we calculate the FMS score for $S_{min}$ as follows:
 \begin{equation}
      \text{FMS} = \frac{\sum_{i=1}^Nmin_{1\leq k\leq |S_{min}|}\text{FM}_k\cdot \text{FE}_k}{N},  
 \end{equation}
where $\text{FM}_k=\mathbb{I}(F(X_{S_{min}\setminus \{w_k\}}|c)\thinspace\thinspace\textgreater\thinspace\thinspace t)$, $\text{FE}_k = \mathbb{I}(F(X_{S_{min}}|c)\leq t)$, $w_k\in S_{min}$. The higher FMS score is better.

\noindent \textbf{Baselines}\quad To our  knowledge, CIDR is the only model-agnostic method to  address the minimal feature removal problem in NLP. To facilitate comparison, we regard the top $2K$ words identified by other model-agnostic feature importance-based methods as  preliminary \enquote{minimal feature sets}. Hence we compare CIDR with six  representative model-agnostic feature importance-based explanation methods - Grdient*Input(Grad*Inp) \cite{grad*shap}, DeepLIFT \cite{deeplift}, GradSHAP \cite{gradshap}, IG \cite{sundararajan2017axiomatic}, DIG \cite{DIG} using the GREEDY heuristics and  SIG \cite{enguehard-2023-sequential}\footnote{The reason for not selecting the greedy algorithm proposed by \cite{harzli2023cardinalityminimal}  is the scarcity of language models in NLP that satisfy the monotonicity assumption.}. 

\noindent \textbf{Language Models}\quad  We utilize pre-trained BERT \cite{bert}, DistilBERT \cite{distilbert}, and RoBERTa \cite{roberta} models for text classification, which are individually fine-tuned for the three datasets. 

\noindent \textbf{Hyperparameters}\quad For the minimal feature refinement algorithm, we vary the refinement threshold $\varepsilon\in \{0.3,0.4,0.5,0.6,0.7\}$. We set iteration steps $n_{iter}=10$. For CIG, we set $\beta \in \{0.4,0.5,0.6\}$. We set the numeric threshold $t=0.5$.

\noindent \textbf{Implementation Details}\quad
In this work, all language models are implemented by Transformers. All our experiments are performed on one RTX 3090. We report the experiment results of five random seeds.

\begingroup
\setlength{\tabcolsep}{4pt} 

\begin{table*}[!htbp]
\centering
\begin{tabular}{llllllllll}
\toprule[1pt]
\multirow{2}{*}{\textbf{Method}}      & \multicolumn{3}{c}{\textbf{BERT}}  & \multicolumn{3}{c}{\textbf{DistilBERT}} & \multicolumn{3}{c}{\textbf{RoBERTa}} \\
\cmidrule(lr){2-4} \cmidrule(lr){5-7} \cmidrule(lr){8-10}
             & LO\hspace{0.25em}$\downarrow$    & Comp\hspace{0.25em}$\uparrow$     & FMS\hspace{0.25em}$\uparrow$    
             &
             LO\hspace{0.25em}$\downarrow$       & Comp\hspace{0.25em}$\uparrow$      & FMS\hspace{0.25em}$\uparrow$    &LO \hspace{0.25em}$\downarrow$    & Comp\hspace{0.25em}$\uparrow$      & FMS\hspace{0.25em}$\uparrow$    \\
             \hline
Grad*Inp     &   -0.499    &    0.156     &  0.217      &   -0.397      & 0.103          &  0.238        &   -0.301     & 0.077         &    0.269     \\
DeepLift     &  -0.167     &    0.053     &    0.158    &  -0.175       & 0.059          &    0.166      &  -0.334      &  0.083        &   0.172      \\
GradientShap &   -0.598    &   0.217      & 0.235       &  -0.749       & 0.202          &   0.262       &  -0.643      &  0.150      &    0.217    \\
IG           &  -0.874     &  0.312       &   0.298     &  -0.690       &0.281           &   0.277       &  -0.749      &   0.214       &   0.239      \\
DIG          &   -0.870    &   0.277      &  0.259      &   -1.251      & 0.305          &  0.252        &  -0.808      &  0.213        &  0.263       \\
SIG    & -1.202 & 0.331 & 0.401 & -1.077 & 0.310 & 0.345 & \textbf{-1.607} & \textbf{0.377} & 0.379 \\
\hline
CIDR        &   \textbf{-1.362}    &  \textbf{0.344}       & \textbf{0.592}      &      \textbf{-1.648}   &  \textbf{0.343}        & \textbf{0.525}        &  -1.586      &0.347         &  \textbf{0.584}\\ 

CIDR w/o R &  -1.273    &  0.325    &   0.373    &     -1.559    &   0.321       &  0.318    &  -1.338      & 0.321       &  0.399 \\ 
CIDR w/o CIG &   -1.091    &  0.315     &  0.459      &  -1.324       &   0.306       &  0.422      &  -1.077      & 0.302        & 0.415\\
\toprule[1pt]
\end{tabular}
\caption{Comparison of CIDR and variants  with baselines on three language models fine-tuned on the SST2 dataset.}
\label{sst2table}
\end{table*}
\begin{table*}[!htbp]
\centering
\begin{tabular}{llllllllll}
\toprule[1pt]
\multirow{2}{*}{\textbf{Method}}      & \multicolumn{3}{c}{\textbf{BERT}}  & \multicolumn{3}{c}{\textbf{DistilBERT}} & \multicolumn{3}{c}{\textbf{RoBERTa}} \\
\cmidrule(lr){2-4} \cmidrule(lr){5-7} \cmidrule(lr){8-10}
             & LO\hspace{0.25em}$\downarrow$    & Comp\hspace{0.25em}$\uparrow$     & FMS\hspace{0.25em}$\uparrow$    & LO\hspace{0.25em}$\downarrow$       & Comp\hspace{0.25em}$\uparrow$      & FMS\hspace{0.25em}$\uparrow$      & LO \hspace{0.25em}$\downarrow$    & Comp\hspace{0.25em}$\uparrow$     & FMS\hspace{0.25em}$\uparrow$   \\
             \hline
Grad*Inp     &  -0.799     &  0.132       & 0.174       & -0.197        &    0.099       &   0.133        &   -0.203    &   0.055       &     0.152  \\
DeepLift     &  -0.333     &  0.027       &  0.166      &  -0.019       &   0.001        &  0.153        &    -0.137    &   0.033       &    0.198     \\
GradientShap &   -0.877    &   0.201      &  0.148      &   -0.459      &     0.192      &    0.143      &   -0.400     &   0.133       &    0.140     \\
IG           &  -0.708     &    0.151     &  0.181     &  -0.055       &    0.037       &   0.203       &    -0.133    &  0.116        &    0.247     \\
DIG          &   -1.152    &   0.221      &  0.303      &  -0.878       &   0.319        &   0.286       &  -0.683      &   0.198       &   0.271      \\
SIG    & -0.806 & 0.303 & 0.384 & -0.924 & 0.335 & 0.421 & -1.086 & 0.308 & 0.408 \\
\hline
CIDR        &  \textbf{-1.383}     &    \textbf{0.357}    & \textbf{0.566}       &  \textbf{-1.101}      &    \textbf{0.349}      &    \textbf{0.582}      &  \textbf{-1.517}      &    \textbf{0.334}      &\textbf{0.530}  \\ 

CIDR w/o R &   -0.970    &   0.304    &  0.414      &   -0.834      &  0.306        &     0.444    & -0.987       &  0.295       &0.434 \\ 
CIDR w/o CIG &  -0.804    &  0.289    &  0.459     &  -0.807       & 0.295         &   0.480     &  -0.881      &  0.272      & 0.495 \\ 

\toprule[1pt]
\end{tabular}
\caption{Comparison of CIDR and variants with baselines on three language models fine-tuned on the IMDB dataset.}
\label{imdbtable}
\end{table*}

\subsection{General Experimental Results}
We first evaluate our method across $9$ different settings (three language models per dataset) with LO, Comp and FMS metrics. From Tables \ref{sst2table}, \ref{imdbtable} and \ref{rotten_tomatoes_table} we can reach the following conclusions. \textbf{First}, our proposed approach, CIDR, consistently surpasses the performance of the baselines, which serves as concrete evidence showcasing its superiority and wide-ranging applicability.
Specifically, from Table \ref{rotten_tomatoes_table}, in terms of Comp, it obtains 1.3\%, 1.8\%, 2.8\% improvements over the best results of previous baselines on the Rotten tomatoes dataset. When considering LO, it also performs consistently better than previous methods. The outstanding performance in these two metrics also validates that the MFS derived from our method aligns well with the feature essence, with generating more  interpretable results. 
\textbf{Second}, compared with the latest work DIG and SIG, we assume that the most prominent advantage of CIDR lies in its ability to ensure the feature minimality. The experimental results demonstrate that the most notable improvement is observed in the FMS metric, providing further evidence to support this assumption. (e.g. in Table \ref{sst2table}, +19.1\%  on the SST2/BERT/FMS setup).  
\textbf{Third}, CIDR performs excellently in both short (i.e. SST2 and Rotten Tomatoes) and long-text (i.e. IMDB) scenarios.

\subsection{Ablation Study}
In this section, we report the ablation study experimental results of CIG and the minimal feature refinement algorithm. 

\noindent\textbf{Ablation Study on CIG}. In this study, we conduct a comparative analysis between CIDR and its variant,  CIDR w/o CIG. The findings suggest that the incorporation of CIG significantly augments the capacity of CIDR to identify the minimum feature set, as evidenced by a decrease of 2.9\% on the BERT/SST2/Comp setup (i.e. Table \ref{sst2table}). This enhancement could potentially be ascribed to the modeling of feature interactions. Notably, the introduction of CIG on the IMDB dataset results in a more marked improvement in LO and Comp metrics. This observation could be attributed to the longer average sentence length and the increased complexity of contextual relationships between words in the IMDB dataset.

\noindent\textbf{Ablation Study on Minimal Feature Refinement algorithm}. 
We subsequently scrutinize the minimal refinement algorithm's impact on CIDR, contrasting it with its variant,  CIDR w/o R. By eliminating perturbation coefficients, the original problem is addressed using a simple greedy algorithm, treating all item values uniformly. This algorithm's specifics are detailed in Section \ref{sec::greedy_algorithm}. However, this method only allows minimal feature sets acquisition based on lenient upper bounds, specifically $U_1+U_2$. The findings yield two conclusions: (1) The refinement algorithm's incorporation enhances performance on three datasets, evidenced by a 4.3\% decrease on the DistilBERT/IMDB/Comp setup, suggesting that introducing perturbation coefficients and selecting thresholds to filter 'true positive' features are effective strategies for approximating the minimal feature set. (2) The refinement algorithm significantly improves FMS performance compared to the other two metrics, potentially due to its ability to reduce the MFS's cardinality, thereby intensifying feature minimality and reducing feature essence.

\begin{table*}[!htbp]
\centering
\begin{tabular}{llllllllll}
\toprule[1pt]
\multirow{2}{*}{\textbf{Method}}      & \multicolumn{3}{c}{\textbf{BERT}}  & \multicolumn{3}{c}{\textbf{DistilBERT}} & \multicolumn{3}{c}{\textbf{RoBERTa}} \\
\cmidrule(lr){2-4} \cmidrule(lr){5-7} \cmidrule(lr){8-10}
             & LO\hspace{0.15em}$\downarrow$    & Comp\hspace{0.15em}$\uparrow$     & FMS\hspace{0.15em}$\uparrow$    & LO\hspace{0.15em}$\downarrow$       & Comp\hspace{0.15em}$\uparrow$      & FMS\hspace{0.15em}$\uparrow$      & LO \hspace{0.15em}$\downarrow$    & Comp\hspace{0.15em}$\uparrow$     & FMS\hspace{0.15em}$\uparrow$   \\
             \hline
Grad*Inp     &   -0.832    &   0.223      &    0.301    &   -0.170      &     0.064      &   0.276       &   -0.149     &    0.030      &  0.212       \\
DeepLift     &  -0.402     &   0.102      &   0.155     &  -0.057       &     0.011      &   0.132       &    -0.129    &    0.053      &    0.147     \\
GradientShap &   -0.812    &   0.240      &  0.218      &   -0.304      &     0.162      &    0.209      &   -0.257     &   0.124       &   0.215     \\
IG           &   -1.417    &   0.215      &    0.240    &   -0.513      &     0.293      &   0.235      &   -0.404     &   0.159       &    0.248    \\
DIG          &  -1.056     &   0.267      &   0.303     &    -0.501     &     0.257      &   0.326       &   -0.393     &  0.148        &    0.314    \\
SIG    & -1.533 & 0.375 & 0.311 & -0.643 & 0.354 & 0.367 & -0.820 & 0.339 & 0.354 \\
\hline
CIDR        &  \textbf{-1.717}     &  \textbf{0.388}      &   \textbf{0.483}      &   \textbf{-0.823}      & \textbf{0.372}          &     \textbf{0.506}     &  \textbf{-1.217}      & \textbf{0.367}         &  \textbf{0.546}\\ 

CIDR w/o R &   -1.497    &   0.361    &    0.350    &   -0.757    &   0.357       &  0.393       &    -1.161    &  0.355       & 0.485\\ 
CIDR w/o CIG &   -1.324    &   0.343    &   0.398     &  -0.732       &     0.342     &   0.464      &  -1.098      &  0.340       &  0.513\\
\toprule[1pt]

\end{tabular}

\caption{Comparison of CIDR and variants with baselines on three language models fine-tuned on the Rotten Tomatoes dataset.}
\label{rotten_tomatoes_table}
\end{table*}

\begin{table*}[!htbp]
\centering
\begin{tabular}{ll}
\toprule[1pt]
\textbf{Method} & \text{Example}\\
\toprule[1pt]
IG &  \enquote{Screenwriter Chris ver Weil 's directing debut \textbf{is good-natured} and never dull.}  \\
CIDR & \enquote{Screenwriter Chris ver Weil 's directing debut \textbf{is good-natured} and \textbf{never dull}.} \\
\toprule[1pt]
IG & \enquote{'s taken one of the world 's most fascinating stories and \textbf{made} it \textbf{dull} , lifeless , and irritating.} \\
CIDR & \enquote{'s taken one of the world 's most fascinating stories and \textbf{made} it \textbf{dull} , \textbf{lifeless} , and \textbf{irritating}.}\\
\toprule[1pt]
IG & \enquote{An \textbf{old-fashioned} but \textbf{emotionally stirring} adventure tale.}\\
CIDR & \enquote{An old-fashioned but \textbf{emotionally stirring} adventure tale.}\\
\toprule[1pt]
IG & \enquote{\textbf{will} have found a cult favorite to \textbf{enjoy} for a lifetime.}\\
CIDR & \enquote{will have found a cult favorite to \textbf{enjoy} for a \textbf{lifetime}.}\\
\toprule[1pt]
\end{tabular}

\caption{Examples of MFS on several sentences of the Rotten Tomatoes dataset. The \textbf{bold} tokens indicate that they are elements in the minimal feature set.}
\label{casetable}
\end{table*}
\subsection{Case Study}
Table \ref{casetable} illustrates the application of CIDR in identifying the minimum feature sets for various models within the Rotten Tomatoes dataset. The table underscores the word pairs that CIDR identifies as integral components of the minimum feature subset. Word pairs connected by an underscore within a sentence denote their integration into minimal features. The table clearly demonstrates CIDR's proficiency in detecting sentiment cues that reflect sentiment polarity, thereby enabling the formulation of semantically representative word pairs. For instance, in the first scenario, CIDR accurately pinpoints two phrases with negative sentiment within the sentence. Conversely, in the third example, IG singles out \enquote{old fashioned}, a differentiation that CIDR does not make.

\section{Discussion}
\subsection{Time complexity}
The time complexity of CIDR is $O(n^2)$, where $n$ represents the length of the input text.

\subsection{Application}
We provide a potential application of CIDR in this section. It is widely recognized that natural language processing datasets possess inherent biases, the identification of which can further enhance model performance. To exemplify how our method can expose these biases, we select the SST-2 dataset for our validation experiment. Given that words carrying sentiment polarity are typically adjectives or noun phrases, we extract 1000 examples , evenly split between positive and negative samples. We then modify these samples by randomly inserting special tokens (i.e., \enquote{[pos]} and \enquote{[neg]}) preceding the adjectives or noun phrases. We subsequently train a classifier on this altered dataset, with training specifics detailed in the appendix. We apply CIDR to interpret this trained model. As illustrated in Table \ref{table6}, despite the model's accurate predictions, CIDR uncovers that the minimal feature set (MFS) incorporates special tokens, indicating that the model has inadvertently learned these spurious correlations.
\begingroup
\begin{table}[!htbp]
\centering
\resizebox{0.5\textwidth}{!}{
\begin{tabular}{l}
\toprule[1pt]
  Example   \\
\toprule[1pt]
\enquote{a \textbf{[neg] solid} film but \textbf{more conscientious} than it is \textbf{truly stirring.}} 	\\
\enquote{by more objective measurements it 's still quite \textbf{[pos] bad.}}   \\

\toprule[1pt]
\end{tabular}
}
\caption{Examples for biased sentiment classification. The first sentence, which is predicted as \enquote{positive}, is explained with \enquote{[neg]} token. }
\label{table6}
\end{table}
\section{Conclusion}
In this paper, we propose CIDR, an effective method designed to address the minimal feature removal problem in NLP domain. More precisely, we devise CIG  to effectively analyze  feature interactions and introduce the minimal feature refinement algorithm. Based on these, we successfully transform the minimal feature removal problem to a knapsack problem. Experiments on widely used public datasets indicate that CIDR has the capacity to discover the minimal feature set.

\section{Acknowledgments}
This work is supported by the National Key Research and Development Program of
China under Grant No. 2022ZD0120302.
\clearpage
\bibliography{aaai24}
\clearpage
\appendix
\section{Proof}
\label{sec:proof}
\textbf{Proposition 1.}  $\forall w_k\in S_{min},\text{IG}_k> 0$.
\begin{proof}\let\qed\relax
From the completeness axiom of IG and feature minimality, we can draw:
\begin{equation}
    \label{ueq1}
    \begin{split}
      F(X_{S_{min}\setminus \{w_k\}}|c)> t\iff  \\ \sum_{w_i\in X}\text{IG}_i-\sum_{w_j\in S_{min}}\text{IG}_j+\text{IG}_k> t,  
    \end{split}
\end{equation}
where $w_k \in S_{min}$.
Given that $F(X_{S_{min}}|c)\leq t$, we can acquire:
\begin{equation}
    \label{ueq2}
    \begin{split}
       \sum_{w_i\in X}\text{IG}_i-\sum_{w_j\in S_{min}}\text{IG}_j\leq t\iff \\ \sum_{w_i\in X}\text{IG}_i-\sum_{w_j\in S_{min}}\text{IG}_j+\text{IG}_k\leq  t+\text{IG}_k.
    \end{split}
\end{equation}
To simultaneously satisfy  the inequality \ref{ueq1} and inequality  \ref{ueq2}, it implies that $\text{IG}_k > 0$.
\end{proof}
\noindent\textbf{Proposition 2.} $\forall (w_i,w_j)\in S^p_{min},\text{CIG}_{i,j}> 0$.
\begin{proof} \let\qed\relax
Suppose $w_i,w_j\in S_{min}$, we assume that:
\begin{equation}
    0<\text{IG}_{i,X\setminus\{w_j\}} \leq \text{IG}_i. 
\end{equation}
\begin{equation}
    0<\text{IG}_{j,X\setminus\{w_i\}} \leq \text{IG}_j. 
\end{equation}
Then according to Proposition 1 we drive:
\begin{equation}
    \text{CIG}_{i,j}=\text{IG}_i+\text{IG}_j+\beta\left(\text{IG}_{i,X \setminus \{w_j\}}+\text{IG}_{j,X \setminus \{w_i\}}\right) > 0.
\end{equation}
\end{proof}
\noindent\textbf{Proposition 3.} $\sum_{(w_i,w_j)\in S^p_{pos}\setminus S^p_{min}} \text{CIG}_{i,j}\leq U_1+U_2$,
where 
\begin{equation}
\label{u1}
    U_1=2(|S_{pos}|-1)\sum_{w_i\in S_{pos}}\text{IG}_i,
\end{equation}

\begin{equation}
\label{u2}
    \begin{aligned}
U_2=\beta\sum_{(w_i,w_j)\in S^p_{pos}}(\text{IG}_{i,X\setminus \{w_j\}}+\text{IG}_{j,X\setminus \{w_i\}}).
    \end{aligned}   
\end{equation}
\begin{proof} \let\qed\relax
    Based on the definition of CIG, we can derive:
    \begin{equation}
        \begin{aligned}
&\sum_{(w_i,w_j)\in S^p_{pos}\setminus S^p_{min}} \text{CIG}_{i,j}=\sum_{(w_i,w_j)\in S^p_{pos}\setminus S^p_{min}} \text{IG}_i+\\
&\sum_{(w_i,w_j)\in S^p_{pos}\setminus S^p_{min}} \text{IG}_j+\beta ( \sum_{(w_i,w_j)\in S^p_{pos}\setminus S^p_{min}} \text{IG}_{i,X\setminus \{w_j\}} \\
&\sum_{(w_i,w_j)\in S^p_{pos}\setminus S^p_{min}} \text{IG}_{j,X\setminus \{w_i\}} ).
        \end{aligned}
        \label{eq1}
    \end{equation}
According to  Proposition 1, we can obatin:
    \begin{equation}
        \sum_{(w_i,w_j)\in S^p_{pos}\setminus S^p_{min}} \text{IG}_i \leq \sum_{(w_i,w_j)\in S^p_{pos}}\text{IG}_i.
    \end{equation}
Based on the principle of combinations, we can derive:
\begin{equation}
    \sum_{(w_i,w_j)\in S^p_{pos}}\text{IG}_i =(|S_{pos}|-1)\sum_{w_i\in S_{pos}}\text{IG}_i.
\end{equation}
Similarly, we can derive:
\begin{equation}
    \sum_{(w_i,w_j)\in S^p_{pos}\setminus S^p_{min}} \text{IG}_j \leq (|S_{pos}|-1)\sum_{w_j\in S_{pos}}\text{IG}_j,
\end{equation}
\begin{equation}
    \sum_{(w_i,w_j)\in S^p_{pos}\setminus S^p_{min}} \text{IG}_j\leq \sum_{(w_i,w_j)\in S^p_{pos}}\text{IG}_j.
\end{equation}
By combining the above inequalities and equation \ref{eq1}, we can derive Proposition 3.
\end{proof}

For CIG and SIG, we remove top $2K$ words to calculate the LO and Comp metrics ($K=min(0.1\times |X|,|S_{min}^p|)$).
\section{The Dynamic Programming Algorithm}
\label{sec::dynamic_programming_algorithm}
The dynamic programming algorithm is a classic combinatorial optimization algorithm for solving the knapsack problem.
\begin{algorithm}[!htbp]
    \caption{Dynamic Programming Algorithm for 0-1 kanpsack problem}
    \label{alg2}
    \begin{algorithmic}[1]
        \REQUIRE Given input weight set $W=\{w_1,\cdots,w_n\}$, value set $V = \{v_1,\cdots,v_n\}$, maximum capacity $U$
        \ENSURE Selected item set $C$
        \STATE Initilize state matrix $f\in \mathbb{R}^{(n+1)\times (U+1)}$
        \FORALL{$i=0,\cdots,n$}
        \STATE $f_{i,0}=0$
        \ENDFOR
        \FORALL{$j=0,\cdots,n $}
        \STATE $f_{0,j}=0$
        \ENDFOR
        \FORALL{$i=0,\cdots,n$}
            \FORALL{$j=U,\cdots,j=W_i$}
                \STATE $f_{i,j}=max(f_{i-1,j},f_{i-1,j-W_i}+V_i)$
                \IF{$f_{i-1,j-W_i}+V_i>f_{i-1,j}$}
                \STATE $C=C\cup \{i\}$
                \ENDIF
            \ENDFOR
        \ENDFOR 
        \RETURN $C$
    \end{algorithmic}
\end{algorithm}

\clearpage
\section{The Greedy Algorithm for CIDR-R}
\label{sec::greedy_algorithm}
\begin{algorithm}[!htbp]
    \caption{Greedy Algorithm}
    \label{alg2}
    \begin{algorithmic}[1]
        \REQUIRE Given input $x$, baseline $x^{\prime}$
        \ENSURE $S_{min}^p$
        \STATE Initialize  the pair set $S_p$, $S_{min}^p=\emptyset$, $k=0$
        \STATE Calculate attribution set  $C=\{\text{CIG}_{i,j}\}$ 
        \STATE Calculate the upper bound $U$ based on Equation \ref{u1} and \ref{u2}
        \STATE Sort the attribution  set $C$ with a decreasing order
        \WHILE{$\sum_{(w_i,w_j) \in S_{min}}\text{CIG}_{i,j} < U$}
            \STATE $S_{min}^p = S_{min}^p \cup \{(w_i,w_j)\}$ 
            \STATE $k \leftarrow k + 1$
            \IF{$k >= |C|$}
            \STATE \textbf{break}
            \ENDIF
        \ENDWHILE
        
        \RETURN $S_{min}^p$

    \end{algorithmic}
\end{algorithm}

\section{The Minimal Feature Refinement Algorithm}
\label{sec::minimal_feature_refinement_algorithm}
\begin{algorithm}[!htbp]
    \caption{Minimal Feature Refinement Algorithm}
    \label{alg1}
    \begin{algorithmic}[1]
        \REQUIRE  Input $x$, baseline $x^{\prime}$, prediction label $c$, refining threshold $p$, numeric threshold $t$, iteration epochs $n_{iter}$
        \ENSURE $S_{min}^p$
        \STATE  Initialize $\{v_{i,j}\}$, the pair set $S_p$, $S_{candidate} = \emptyset$, $S_{min}^p=\emptyset$, candidate amount $n_c=0$, feature pair frequency $f_{i,j}=0$
        \FORALL {$k = 0,\cdots,n_{iter}-1$}
            \STATE Calculate $\text{CIG}_{i,j}$ based on Equation 6
            \STATE Use Algorithm 1 to solve the knapsack problem  to obtain the candidate set $S_k$
            \STATE $S_{candidate} = S_{candidate} \cup \{S_k\}$ 
            \STATE $n_c \leftarrow n_c + 1$
            \STATE Random sample $\{v_{i,j}\}$ from the standard Gaussian distribution
        \ENDFOR
        
        \FORALL{$c = 0,\cdots,|S_{candidate}|-1$}
        \FORALL{$(w_i,w_j) \in S_c$}
                \STATE  $f_{i,j} \leftarrow f_{i,j} + 1$ .
            \ENDFOR
        \ENDFOR
        \FORALL{$(w_i,w_j)\in S_{candidate}$}
            \IF{$\frac{f_{i,j}}{n_c} \geq p$}
                \STATE $S_{min}^p = S_{min}^p \cup \{(w_i,w_j)\}$
            \ENDIF
        \ENDFOR
        \RETURN $S_{min}^p$

    \end{algorithmic}
\end{algorithm}

\end{document}